\documentclass[10pt,twocolumn,letterpaper]{article}

\usepackage[pagenumbers]{cvpr} 

\usepackage{graphicx}
\usepackage{amsmath}
\usepackage{amssymb}
\usepackage{booktabs}
\usepackage{lipsum}  
\usepackage[utf8x]{inputenc} 
\usepackage[pagebackref,breaklinks,colorlinks]{hyperref}

\usepackage[capitalize]{cleveref}
\crefname{section}{Sec.}{Secs.}
\Crefname{section}{Section}{Sections}
\Crefname{table}{Table}{Tables}
\crefname{table}{Tab.}{Tabs.}

\def\cvprPaperID{*****} 
\def\confName{CVPR}
\def\confYear{2022}

\usepackage{tabularx}
\usepackage{booktabs}
\usepackage{bm}
\usepackage[free-standing-units=true]{siunitx}
\usepackage{xcolor}
\usepackage{multirow}
\usepackage[normalem]{ulem}
\useunder{\uline}{\ul}{}
\usepackage[utf8x]{inputenc} 
\usepackage[pagebackref,breaklinks,colorlinks]{hyperref}
\usepackage[capitalize]{cleveref}

\pdfoutput=1
\begin{document}
\title{SSL-Lanes: Self-Supervised Learning for Motion Forecasting in Autonomous Driving
\vspace{-0.4cm}}

\author{
Prarthana Bhattacharyya \hspace{0.5cm}  Chengjie Huang \hspace{0.5cm}  Krzysztof Czarnecki\\
University of Waterloo, Canada\\
{\tt\small \{p6bhatta, c.huang, k2czarne\}@uwaterloo.ca}
}

\twocolumn[{%
\renewcommand\twocolumn[1][]{#1}%
\maketitle
\begin{center}
\vspace{-0.4cm}
    \centering
    \captionsetup{type=figure}
    \includegraphics[width=\textwidth]{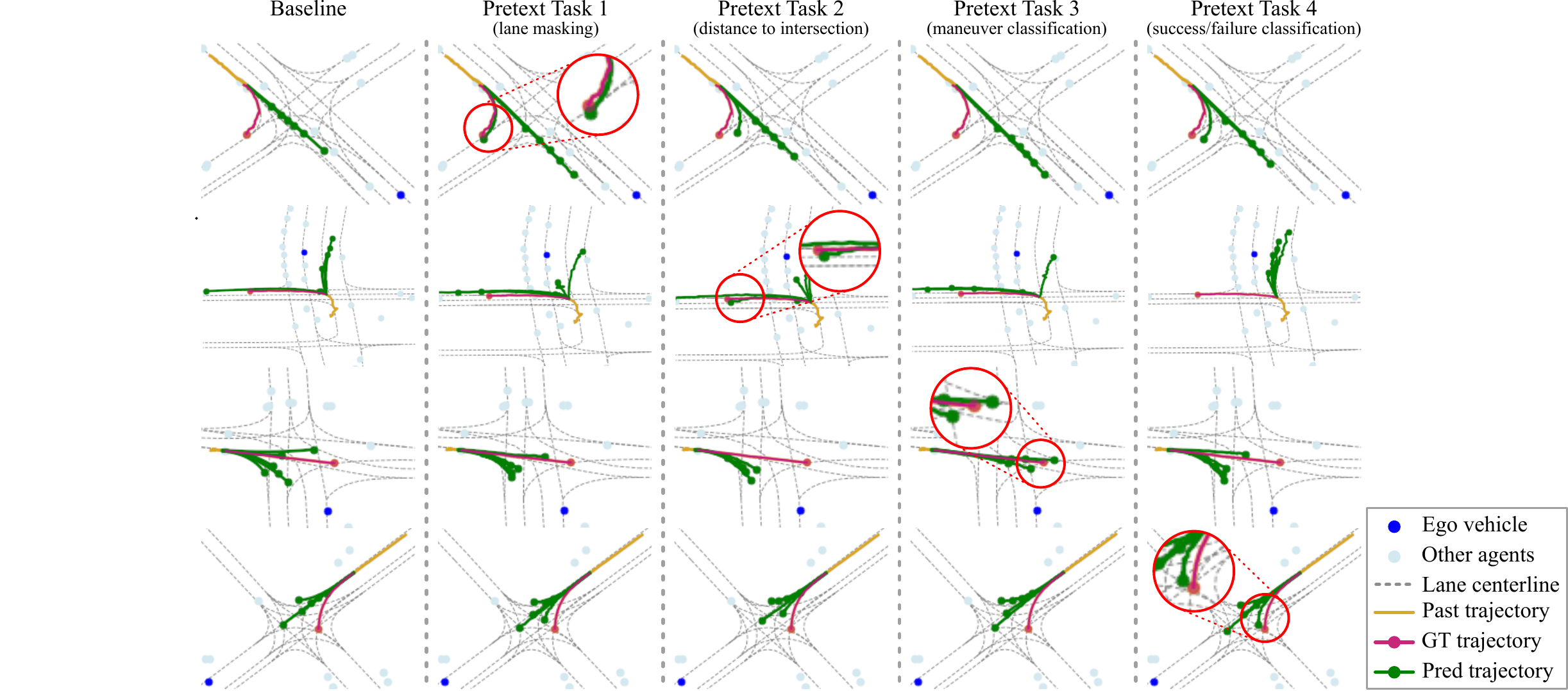}
    \caption{Motion forecasting on Argoverse \cite{Argoverse} validation. We show four challenging scenarios at intersections. The baseline \cite{LaneGCN} misses all the predictions. In the first row, our proposed lane masking successfully captures the right-turn. For the second row, predicting distance to intersection helps the most in capturing the left turn. In the third row, acceleration at an intersection is best captured by the model that is made to classify maneuvers of traffic agents. Finally, in the fourth row, classifying successful final goal states is the most effective at capturing the left turn. These tasks are trained with pseudo-labels which are obtained for free from data. Please refer to \cref{qualitative} for details.}
   \label{fig:qualitative}
\end{center}%
}]

\begin{abstract}
\vspace{-0.4cm}
Self-supervised learning (SSL) is an emerging technique that has been successfully employed to train convolutional neural networks (CNNs) and graph neural networks (GNNs) for more transferable, generalizable, and robust representation learning. However its potential in motion forecasting for autonomous driving has rarely been explored. In this study, we report the first systematic exploration and assessment of incorporating self-supervision into motion forecasting. We first propose to investigate four novel self-supervised learning tasks for motion forecasting with theoretical rationale and quantitative and qualitative comparisons on the challenging large-scale Argoverse dataset. Secondly, we point out that our auxiliary SSL-based learning setup not only outperforms forecasting methods which use transformers, complicated fusion mechanisms and sophisticated online dense goal candidate optimization algorithms in terms of performance accuracy, but also has low inference time and architectural complexity. Lastly, we conduct several experiments to understand why SSL improves motion forecasting. Code is open-sourced  at \url{https://github.com/AutoVision-cloud/SSL-Lanes}.
\vspace{-0.4cm}
\end{abstract}

\section{Introduction}
\label{sec:intro}
Motion forecasting in a real-world urban environment is an important task for autonomous robots. It involves predicting the future trajectories of traffic agents including vehicles and pedestrians. This is absolutely crucial in the self-driving domain for safe, comfortable and efficient operation. However, this is a very challenging problem. Difficulties include inherent stochasticity and multimodality of driving behaviors, and that future motion can involve complicated maneuvers such as yielding, nudging, lane-changing, turning and acceleration or deceleration.

The motion prediction task has traditionally been based on kinematic constraints and road map information with handcrafted rules. These approaches however fail to capture long-term behavior and interactions with map structure and other traffic agents in complex scenarios. Tremendous progress has been made with data-driven methods in motion forecasting \cite{Multipath, Jean, TNT, DenseTnT, WIMP, SceneTransformer, mmtransformer, Intentnet}. Recent methods use a vector representation for HD maps and agent trajectories, including approaches like Lane-GCN \cite{LaneGCN}, Lane-RCNN \cite{Lanercnn}, Vector-Net \cite{Vectornet}, TNT \cite{TNT} and Dense-TNT \cite{DenseTnT}. More recently, the enormous success of transformers \cite{attentionisallyouneed} has been leveraged for forecasting in mm-Transformer \cite{mmtransformer}, Scene transformer \cite{SceneTransformer}, Multimodal transformer \cite{Waypoint} and Latent Variable Sequential Transformers \cite{LVT}. Most of these methods however are extremely complex in terms of architecture and have low inference speeds, which makes them unsuitable for real-world settings. 

In this work, we extend ideas from self-supervised learning (SSL) to the motion forecasting task. Self-supervision has seen huge interest in both natural language processing and computer vision \cite{DINO} to make use of freely available data without the need for annotations. It aims to assist the model to learn more transferable and generalized representation from pseudo-labels via pretext tasks. Given the recent success of self-supervision with CNNs, transformers, and GNNs, we are naturally motivated to ask the question: \textit{Can self-supervised learning 
improve accuracy and generalizability of motion forecasting, without sacrificing inference speed or architectural simplicity?} 


\vspace{0.3cm}
\textbf{Contributions:} Our work, SSL-Lanes, presents the first systematic study on how to incorporate self-supervision in a standard data-driven motion forecasting model. Our contributions are:
\vspace{-0.2cm}
\begin{itemize}
\item We demonstrate the effectiveness of incorporating self-supervised learning in motion forecasting. Since this does not add extra parameters or compute during inference, SSL-Lanes achieves the best accuracy-simplicity-efficiency trade-off on the challenging large-scale Argoverse \cite{Argoverse} benchmark.
\vspace{-0.2cm}
\item We propose four self-supervised tasks based on the nature of the motion forecasting problem. The key idea is to leverage easily accessible map/agent-level information to define domain-specific pretext tasks that encourage the standard model to capture more superior and generalizable representations for forecasting, in comparison to pure supervised learning.
\vspace{-0.1cm}
\item We further design experiments to explore why forecasting benefits from SSL. We provide extensive results to hypothesize that SSL-Lanes learns richer features from the SSL training as compared to a model trained with vanilla supervised learning.
\end{itemize}

\section{Related Work}
\label{sec:relatedwork}
\textbf{Motion Forecasting:}
Traditional methods for motion forecasting primarily use Kalman filtering \cite{kalman1960} with a prior from HD-maps to predict future motion states \cite{physics+maneuver, motionmodel+manuever}. With the huge success of deep learning, recent works use data-driven approaches 
for motion forecasting. These methods explore different architectures involving rasterized images and CNNs \cite{Multipath, chaueffeurnet, Covernet}, vectorized representations and GNNs \cite{Vectornet, Lanercnn, SocialSTGCNN, Jean, WIMP}, point-cloud representations \cite{TPCN}, transformers \cite{SceneTransformer, mmtransformer, LVT, Waypoint} and sophisticated fusion mechanisms \cite{LaneGCN}, to generate features that predict final output trajectories. While the focus of these works is to find more effective ways of feature extraction from HD-maps and interacting agents, they need huge model capacity, heavy parameterization, and extensive augmentations or large amounts of data to converge to a general solution. 
Other works \cite{TNT, Intentnet, PRIME, neuralmotionplanner} build on them to incorporate prior knowledge in the form of predefined candidate trajectories from sampling or clustering strategies from training data. However the disadvantage of these methods is that their performance is highly related to the quality of the trajectory proposals, which becomes an extra dependency. End-to-end solutions for optimizing end-points of these candidates trajectories are proposed by Dense-TNT \cite{DenseTnT} and HOME \cite{HOME}. Dense-TNT has state-of-the-art accuracy with a reasonable parameter budget, but its online dense goal candidate optimization strategy is computationally very expensive, which is unrealistic for real-time operations like autonomous driving. Lately, ensembling techniques like MultiPath++ \cite{multipath++} and DCMS \cite{DCMS} have been proposed. While they have high forecasting performance, a major disadvantage is their high memory cost for training and heavy computational cost at inference. 

\par \textbf{Self-supervised Learning:}
SSL is a rapidly emerging learning framework that generates additional supervised signals to train deep learning models through carefully designed pretext tasks. In the image domain, various self-supervised learning techniques have been developed for learning high-level image representations, including predicting the relative locations of image patches \cite{DoerschGE15}, jigsaw puzzle \cite{jigsaw}, image rotation \cite{imagerotation}, image clustering \cite{clustering}, image inpainting \cite{inpainting}, image colorization \cite{colorization} and segmentation prediction \cite{segmentation}. In the domain of graphs and graph neural networks, pretext tasks include graph partitioning, node clustering, context prediction and graph completion \cite{SSL_GCN, SSL_on_Graphs, Survey_GraphSSL, PretrainGNN}. To the best of our knowledge, this is the first principled approach that explores motion forecasting for autonomous driving with self-supervision.

\begin{figure*}[t]
  \centering
   \includegraphics[width=0.9\textwidth,height=7cm]{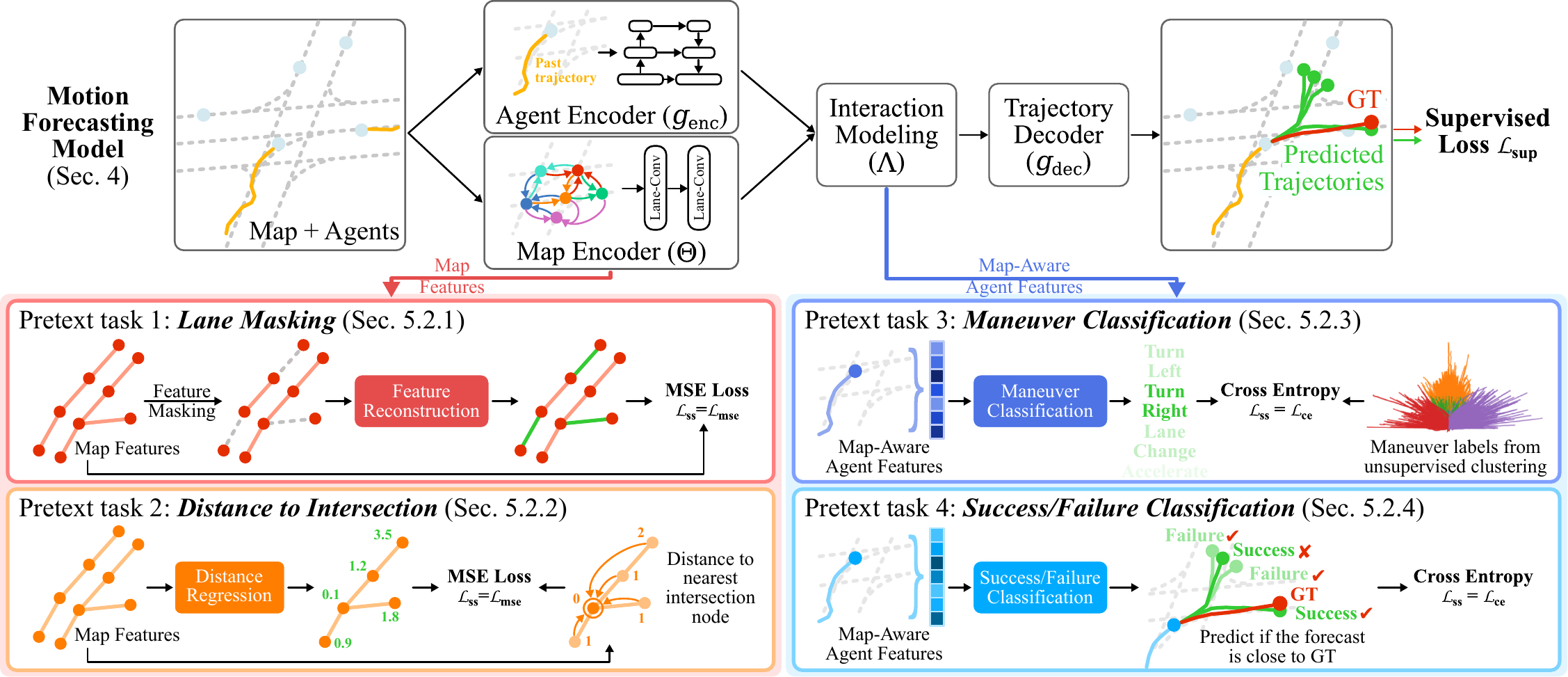}
 \caption{Illustration  of the overall SSL-Lanes framework for self-supervision on motion forecasting through joint training. SSL-Lanes improves upon a standard-motion forecasting baseline, that consists of an agent encoder, map encoder, interaction model and a trajectory decoder, trained using a supervised loss $\mathcal{L}_{sup}$. SSL-Lanes proposes four pretext tasks: (1) Lane Masking: which recovers feature information from the perturbed lane graphs. (2) Distance to Intersection: which predicts the distance (in terms of shortest path length) from all lane nodes to intersection nodes. (3) Maneuver Classification: predicts the form of a ‘maneuver’ the agent-of-interest intends to execute (4) Success/Failure Classification: which trains an agent specialized at achieving end-point goals.}
   \label{methods}
   \vspace{-0.3cm}
\end{figure*}
\section{Problem Formulation}
\label{sec:problem}
We are given the past motion of $N$ actors. The $i$-th actor is denoted as a set of its center locations over the past $L$ time-steps. We pre-process it to represent each trajectory as a sequence of displacements $\mathcal{P}_i = \{\Delta \boldsymbol{p}_i^{−L+1}, . . . , \Delta \boldsymbol{p}_i^{−1}, \Delta \boldsymbol{p}_i^{0}\}$, where $\boldsymbol{p}_i^l$ is the 2D displacement from time step $l - 1$ to $l$. We are also given a high-definition (HD) map, which contains lanes and semantic attributes. Each lane is composed of several consecutive lane nodes, with a total of $M$ nodes. $\boldsymbol{X} \in \mathbb{R}^{M \times F}$ denotes the lane node feature matrix, where $\boldsymbol{x}_j=\boldsymbol{X}[j, :]^T$ is the $F$-dimensional lane node vector. Following the connections between lane centerlines (i.e., predecessor, successor, left neighbour and right neighbour), we represent the connectivity among the lane nodes with four adjacency matrices $\{\boldsymbol{A}_f\}_{f\in \{ \text{pre,} \text{suc,} \text{left,} \text{right} \}}$, with $\boldsymbol{A}_f \in \mathbb{R}^{M \times M}$. This implies that if $\boldsymbol{A}_{f,gh} = 1$, then node $h$ is an $f$-type neighbor of node $g$. Our goal is to forecast the future motions of all actors in the scene $\mathcal{O}^{1:T}_{\text{GT}} = \{(x^1_i , y^1_i ),..., (x^T_i , y^T_i ) | i = 1, ..., N\}$, where $T$ is our prediction horizon. 

\section{Background}
\label{sec:background}
In this section, we first briefly introduce a standard data-driven motion forecasting framework.
\vspace{0.1cm}
\par \textit{Feature Encoding:} We first encode the agent and map inputs similar to Lane-GCN \cite{LaneGCN}. The agent encoder includes a 1D convolution with a feature pyramid network, parameterized by $g_{\text{enc}}$, as given by \cref{eq:1}. For map-encoding, we adopt two Lane-Conv residual blocks, parameterized by $ \boldsymbol{\mathrm{\Theta}} = \{\boldsymbol{W}_0, \boldsymbol{W}_{\text{left}}, \boldsymbol{W}_{\text{right}}, \boldsymbol{W}_{\text{pre}, k}, \boldsymbol{W}_{\text{suc}, k}\}$, where $k\in\{1,2,4,8,16,32\}$,  as given by \cref{eq:2}.
\begin{equation}\label{eq:1}
\small
\hat{\boldsymbol{p}}_i = g_{\text{enc}}(\mathcal{P}_i) 
\normalsize
\end{equation}
\begin{equation}\label{eq:2}
\small
\resizebox{\hsize}{!}{$
\boldsymbol{Y} = \boldsymbol{X} \boldsymbol{W}_0 + \sum\limits_{j \in \{\text{left}, \text{right}\}} {\boldsymbol{A}_j\boldsymbol{X}\boldsymbol{W}_{j}} + \sum\limits_{k}{\boldsymbol{A}_{\text{pre}}^{k}\boldsymbol{X}\boldsymbol{W}_{\text{pre}, k} + \boldsymbol{A}_{\text{suc}}^{k}\boldsymbol{X}\boldsymbol{W}_{\text{suc}, k}}
$}
\end{equation} 
\par \textit{Modeling Interactions:} Since the behavior of agents depends on map topology and social consistency, each encoded agent $i$ subsequently aggregates context from the surrounding map features and its neighboring agent features, via spatial attention \cite{Vaswani} as given by \cref{eq:3}:
\begin{equation}\label{eq:3}
\small
\begin{split}
   \boldsymbol{\tilde{p}}_i = \boldsymbol{\hat{p}}_i\boldsymbol{W}_\text{M2A} + \sum_j{\phi(\text{concat}(\boldsymbol{\hat{p}}_i, \Delta_{i, j} , \boldsymbol{y}_j)\boldsymbol{W}_1)\boldsymbol{W}_2} 
\\
   \boldsymbol{\acute{p}}_i = \boldsymbol{\tilde{p}}_i \boldsymbol{W}_\text{A2A} + \sum_j{\phi(\text{concat}(\boldsymbol{\tilde{p}}_i, \Delta_{i, j} , \boldsymbol{\tilde{p}}_j)\boldsymbol{W}_3)\boldsymbol{W}_4} 
\end{split}
\end{equation}
Here, $\boldsymbol{y}_j$ is the feature of the $j$-th node, $\boldsymbol{\hat{p}}_i$ is the feature of the $i$-th agent, $\phi$ the composition of layer normalization and ReLU, and $\Delta_{ij} = \text{MLP}(\boldsymbol{v}_j −\boldsymbol{v}_i)$, where $\boldsymbol{v}$ denotes the $(x, y)$ 2-D BEV location of the agent or the lane node. The parameters for map and agent feature aggregation is represented by $\boldsymbol{\Lambda} = \{ \boldsymbol{W}_\text{M2A}, \boldsymbol{W}_1, \boldsymbol{W}_2, \boldsymbol{W}_\text{A2A}, \boldsymbol{W}_3, \boldsymbol{W}_4 \}$.
\vspace{0.1cm}
\par \textit{Trajectory Prediction:} Finally, we decode the future trajectories from the features $\boldsymbol{\acute{p}}_i$
corresponding to the agents of interest as given by:
$\mathcal{O}^{1:T}_{\text{pred}} = \{ g_{\text{dec}}(\boldsymbol{\acute{p}}_i) | i = 1, ..., N\}$, where $g_{\text{dec}}$ is the parameterized trajectory decoder. 
The parameters for the motion forecasting model are learned by minimizing the supervised loss ($\mathcal{L}_\text{sup}$) calculated between the predicted output and the ground-truth future trajectories ($\mathcal{O}^{1:T}_{\text{GT}}$), as given by \cref{eq:4}:
\begin{equation} \label{eq:4}
\small
g_{\text{enc}}^{\star}, \boldsymbol{\mathrm{\Theta}}^{\star}, \boldsymbol{\Lambda}^{\star}, g_{\text{dec}}^{\star} =  \mathop{\text{arg min}}_{g_{\text{enc}}, \boldsymbol{\mathrm{\Theta}}, \boldsymbol{\Lambda}, g_{\text{dec}}} \mathcal{L}_{\text{sup}}(\mathcal{O}^{1:T}_{\text{pred}}, \mathcal{O}^{1:T}_{\text{GT}})
\end{equation}

        

\begin{table*}[]
\small
\centering
\begin{tabular}{r|l|l|l}
\toprule
\textbf{SSL Task}              & \textbf{Property Level}                                                             & \textbf{Primary Assumption} & \textbf{Type}                        \\ \hline
Lane-Masking                      & \multirow{2}{*}{Map features}                                                       & Local map structure         & Aux. auto-encoder                    \\ \cline{1-1} \cline{3-4} 
Distance to Intersection       &                                                                                     & Global map structure        & Aux. regression                      \\ \hline
Maneuver Classification        & \multirow{2}{*}{\begin{tabular}[c]{@{}l@{}}Map-aware\\ agent features\end{tabular}} & Agent feature similarity    & \multirow{2}{*}{Aux. classification} \\ \cline{1-1} \cline{3-3}
Success/Failure Classification &                                                                                     & Distance to success state   &                                      \\ \bottomrule
\end{tabular}%
\caption{Overview of our proposed self-supervised (SSL) tasks}
\label{tab:pretext-tasks}
\vspace{-0.16in}
\end{table*}

\section{SSL-Lanes}
\label{sec:method}
The goal of our proposed SSL-Lanes framework is to improve the performance of the primary motion forecasting baseline by learning simultaneously with various self-supervised tasks. \cref{methods} shows the pipeline of our proposed approach, and \cref{tab:pretext-tasks} summarizes the self-supervised tasks.
\subsection{Self-Supervision meets Motion Forecasting}
Before we discuss designing pretext tasks to generate self-supervisory signals, we consider a scheme that will allow combined training for self-supervised pretext tasks and our standard framework.
\vspace{0.1cm}
\par \textbf{How to combine motion forecasting and SSL?} Self-supervision can be combined with motion forecasting in various ways. In one scheme we could pre-train the forecasting encoder with pretext tasks (which can be viewed as an initialization for the encoder’s parameters) and then fine-tune the pre-trained encoder with a downstream decoder as given by \cref{eq:4}. In another scheme, we could choose to freeze the encoder and only train the decoder. In a third scheme, we could optimize our pretext task and primary task \textit{jointly}, as a kind of multi-task learning setup. Inspired by relevant discussions in GNNs, we choose the third-scheme, i.e., multi-task learning, which is the most general framework among the three and is also experimentally verified to be the most effective \cite{SSL_GCN, SSL_on_Graphs}.
\vspace{0.1cm}
\par \textbf{Joint Training:} Considering our motion forecasting task and a self-supervised task, the output and the training process can be formulated as:
\begin{equation} \label{eq:5}
\boldsymbol{\mathrm{\Psi}}^{\star}, \boldsymbol{\mathrm{\Omega}}^{\star}, \boldsymbol{\mathrm{\Theta}}_{\text{ss}}^{\star} = 
\mathop{\text{arg min}}_{\boldsymbol{\mathrm{\Psi}}, \boldsymbol{\mathrm{\Omega}}, \boldsymbol{\mathrm{\Theta}}_{\text{ss}}} \quad \alpha_1  \mathcal{L}_{\text{sup}}(\boldsymbol{\mathrm{\Psi}}, \boldsymbol{\mathrm{\Omega}}) + \alpha_2 \mathcal{L}_{\text{ss}}(\boldsymbol{\mathrm{\Psi}}, \boldsymbol{\mathrm{\Theta}}_{\text{ss}})
\end{equation}
where, $\mathcal{L}_{\text{ss}}(\cdot, \cdot)$ is the loss function of the self-supervised task, $\boldsymbol{\mathrm{\Theta}}_{\text{ss}}$ is the corresponding linear transformation parameter, and $\alpha_1, \alpha_2 \in \mathrm{R}_{>0}$ are the weights for the supervised and self-supervised losses. If the pretext task only focuses on the map encoder, then $\boldsymbol{\mathrm{\Psi}} = \{\boldsymbol{\mathrm{\Theta}}\}$ and $\boldsymbol{\mathrm{\Omega}} = \{ {g_{\text{enc}}, \boldsymbol{\Lambda}},  g_{\text{dec}} \}$. Otherwise, $\boldsymbol{\mathrm{\Psi}} = \{ {g_{\text{enc}}, \boldsymbol{\mathrm{\Theta}}, \boldsymbol{\Lambda}} \}$ and $\boldsymbol{\mathrm{\Omega}}=\{ g_{\text{dec}} \}$. Henceforth, we also define the following representations. We will represent the primary task encoder as function $f_{\boldsymbol{\mathrm{\Psi}}}$, parameterized by $\boldsymbol{\mathrm{\Psi}}$. Furthermore, given a pretext task, which we will design in the next section, the pretext decoder $p_{\boldsymbol{\mathrm{\Theta}}_{\text{ss}}}$ is a function that predicts pseudo-labels and is parameterized by $\boldsymbol{\mathrm{\Theta}}_{\text{ss}}$.
\par \textbf{Benefit of SSL-Lanes:} In \cref{eq:5}, the self-supervised task as a regularization term throughout network training. It acts as the regularizer learned from unlabeled data under the minor guidance of human prior (design of pretext task). Therefore, a properly designed task would introduce data-driven prior knowledge that improves model generalizability.
\subsection{Pretext tasks for Motion Forecasting}
At the core of our SSL-Lanes approach is defining pretext tasks based upon self-supervised information from the underlying map structure \emph{and} the overall temporal prediction problem itself. 
Our proposed prediction-specific self-supervised tasks are summarized in \cref{tab:pretext-tasks}, and assign different pseudo-labels from unannotated data to solve \cref{eq:5}. Our core approach is simple in contrast to state-of-the-art that rely on complex encoding architectures \cite{Lanercnn, mmtransformer, SceneTransformer, WIMP, LaneGCN, Waypoint, TNT}, ensembling forecasting heads \cite{DCMS, multipath++}, involved final goal-set optimization algorithms \cite{DenseTnT, PRIME} or heavy fusion mechanisms \cite{LaneGCN}, to improve prediction performance. 

\begin{table*}[]
\small
\centering
\begin{tabular}{@{}r|ccc|ccc@{}}
\toprule
\textbf{Method}                & \textbf{$\textrm{minADE}_1$} & \textbf{$\textrm{minFDE}_1$} & \textbf{$\textrm{MR}_1$} & \textbf{$\textrm{minADE}_6$} & \textbf{$\textrm{minFDE}_6$} & \textbf{$\textrm{MR}_6$} \\ \midrule
Baseline                       & 1.42                         & 3.18                         & 51.35                    & 0.73                         & 1.12                         & 11.07                    \\ \midrule 
Lane-Masking                     & 1.36                         & 2.96                         & 49.45                    & {\ul \textbf{0.70}}          & 1.02                         & 8.82                     \\
Distance to Intersection       & 1.38                         & 3.02                         & 49.53                    & 0.71                         & 1.04                         & 8.93                     \\
Maneuver Classification        & {\ul \textbf{1.33}}          & {\ul \textbf{2.90}}          & 49.26                    & 0.72                         & 1.05                         & 9.36                     \\
Success/Failure Classification & 1.35                         & 2.93                         & {\ul \textbf{48.54}}     & {\ul \textbf{0.70}}          & {\ul \textbf{1.01}}          & {\ul \textbf{8.59}}      \\ \bottomrule
\end{tabular}%
\vspace{0.05in}
\caption{Motion forecasting performance on Argoverse validation with our proposed pretext tasks}
\label{tab:ablation}
\vspace{-0.1in}
\end{table*}

\vspace{-0.12in}
\subsubsection{Lane-Masking}
\quad \textbf{Motivation:} The goal of the \textit{Lane-Masking} pretext task is to encourage the map encoder $\boldsymbol{\mathrm{\Psi}} = \{\boldsymbol{\mathrm{\Theta}}\}$ to learn local structure information in addition to the forecasting task that is being optimized. In this task, we learn by recovering feature information from the perturbed lane graphs. VectorNet \cite{Vectornet} is the only other motion forecasting work that proposes to randomly mask out the input node features belonging to either scene context or agent trajectories, and ask the model to reconstruct the masked features. Their intuition is to encourage the graph networks to better capture the interactions between agent dynamics and scene context. However, our motivation differs from VectorNet in two respects: (a) We propose to use masking to learn local map-structure better, as opposed to learning interactions between map and the agent. This is an easier optimization task, and we outperform VectorNet. (b) A lane is made up of several nodes. We propose to randomly mask out a certain percentage of each lane. This is a much stronger prior as compared to randomly masking out \textit{any} node and ensures that the model pays attention to all parts of the map. 
\par \textbf{Formulation:} Formally, we randomly mask (i.e., set equal to zero) the features of $m_a$ percent of nodes per lane and then ask the self-supervised decoder to reconstruct these
features. 
\begin{equation} \label{eq:6}
\small
\boldsymbol{\mathrm{\Psi}}^{\star}, \boldsymbol{\mathrm{\Theta}}_{\text{ss}}^{\star} = 
\mathop{\text{arg min}}_{\boldsymbol{\mathrm{\Psi}}, \boldsymbol{\mathrm{\Theta}}_{\text{ss}}}  \frac{1}{m_a} \sum_{i=1}^{m_a} \mathcal{L}_{\text{mse}}\Big(p_{\boldsymbol{\mathrm{\Theta}}_{\text{ss}}}([f_{\boldsymbol{\mathrm{\Psi}}}(\Tilde{\boldsymbol{X}}, \boldsymbol{A}_f)]_{\boldsymbol{v}_i}), \boldsymbol{X}_i \Big)
\end{equation}
Here, $\Tilde{\boldsymbol{X}}$ is the node feature matrix corrupted with random masking, i.e., some rows of $\boldsymbol{X}$ corresponding to nodes $\boldsymbol{v}_i$ are set to zero.
$p_{\boldsymbol{\mathrm{\Theta}}_{\text{ss}}}$ is a fully connected network that maps the representations to the reconstructed features. $\mathcal{L}_{\text{mse}}$ is the mean squared error (MSE) loss function penalizing the distance between the reconstructed map features $p_{\boldsymbol{\mathrm{\Theta}}_{\text{ss}}}([f_{\boldsymbol{\mathrm{\Psi}}}(\Tilde{\boldsymbol{X}}, \boldsymbol{A}_f)]_{\boldsymbol{v}_i})$ for node $\boldsymbol{v}_i$ and its GT features $\boldsymbol{X}_i$.
\vspace{0.15cm}
\par \textbf{Benefit of Lane-Masking:} Since Argoverse \cite{Argoverse} has imbalanced data with respect to maneuvers, there are cases when right/left turns, lane-changes, acceleration/deceleration are missed by the baseline even with multi-modal predictions. We hypothesize that stronger map-features can help the multi-modal prediction header to infer that some of the predictions should also be aligned with map topology. For example, even if an agent is likely to go straight at an intersection, some of the possible futures should also cover acceleration/deceleration or right/left turns guided by the local map structure.  
\vspace{-0.3cm}
\subsubsection{Distance to Intersection}
\quad \textbf{Motivation:} The Lane-Masking pretext task is from a local structure perspective based on masking and trying to predict local attributes of the vectorized HD-map. We further develop the \textit{Distance-to-Intersection} pretext task to guide the map-encoder, $\boldsymbol{\mathrm{\Psi}} = \{\boldsymbol{\mathrm{\Theta}}\}$, to maintain global topology information by predicting the distance (in terms of shortest path length) from all lane nodes to intersection nodes. Datasets like Argoverse \cite{Argoverse} provide lane attributes which describe whether a lane node is located within an intersection. This will force the representations to learn a global positioning vector of each of the lane nodes.
\par \textbf{Formulation:} We aim to regress the distances from each lane node to pre-labeled intersection nodes annotated as part of the dataset. Given $K$ labeled intersection nodes $\mathcal{V}_{\text{intersection}} = \{\boldsymbol{v}_{\text{intersection}, k}|k=1,...K\}$, we first generate reliable pseudo labels using breadth-first search (BFS). Specifically, BFS calculates the shortest distance $d_i \in \mathbb{R}$ for every lane node $\boldsymbol{v}_i$ from the given set $\mathcal{V}_{\text{intersection}}$.  The target of this task is to predict the pseudo-labeled distances using a pretext decoder. If $p_{\boldsymbol{\mathrm{\Theta}}_{\text{ss}}}([f_{\boldsymbol{\mathrm{\Psi}}}(\boldsymbol{X}, \boldsymbol{A}_f)]_{\boldsymbol{v}_i})$ is the prediction of node $\boldsymbol{v}_i$, and $\mathcal{L}_{\text{mse}}$ is the mean-squared error loss function for regression, then the loss formulation for this SSL pretext task is as follows:
\begin{equation} \label{eq:7}
\boldsymbol{\mathrm{\Psi}}^{\star}, \boldsymbol{\mathrm{\Theta}}_{\text{ss}}^{\star} = 
\mathop{\text{arg min}}_{\boldsymbol{\mathrm{\Psi}}, \boldsymbol{\mathrm{\Theta}}_{\text{ss}}}  \frac{1}{M} \sum_{i=1}^M \mathcal{L}_{\text{mse}}\Big(p_{\boldsymbol{\mathrm{\Theta}}_{\text{ss}}}([f_{\boldsymbol{\mathrm{\Psi}}}(\boldsymbol{X}, \boldsymbol{A}_f)]_{\boldsymbol{v}_i}), d_i\Big)
\end{equation}
\par \textbf{Benefit of Distance to Intersection Task:} We hypothesize that since change of speed, acceleration, primary direction of movement etc. for an agent can change far more dramatically as an agent approaches or moves away from an intersection, it is beneficial to explicitly incentivize the model to pick up the geometric structure near an intersection and compress the space of possible map-feature encoders, thereby effectively simplifying inference. We also expect this to improve drivable area compliance nearby an intersection, which is often a problem for current motion forecasting models. 
\vspace{-0.3cm}

\begin{table*}[]
\small
\centering
\resizebox{\textwidth}{!}{%
\begin{tabular}{@{}c|ccc|ccc|c@{}}
\toprule
\textbf{Method}                                                                 & \textbf{$\textrm{minADE}_1$} & \textbf{$\textrm{minFDE}_1$} & \textbf{$\textrm{MR}_1$} & \textbf{$\textrm{minADE}_6$} & \textbf{$\textrm{minFDE}_6$} & \textbf{$\textrm{MR}_6$} & \textbf{$\textrm{b-FDE}_6$} \\ \midrule
NN + Map \cite{Argoverse}                                                       & 3.65                         & 8.12                         & 94.0                     & 2.08                         & 4.02                         & 58.0                     & -                           \\
Jean \cite{Jean}                                                                & 1.74                         & 4.24                         & 68.56                    & 0.98                         & 1.42                         & 13.08                    & 2.12                        \\
Lane-GCN \cite{LaneGCN}                                                         & 1.71                         & 3.78                         & 58.77                    & 0.87                         & 1.36                         & 16.20                    & 2.05                        \\
LaneRCNN \cite{Lanercnn}                                                        & {\ul 1.68}       & 3.69                         & {\ul 56.85}  & 0.90                         & 1.45                         & {\ul 12.32}  & 2.15                        \\
TNT \cite{TNT}                                                                  & 1.77                         & 3.91                         & 59.70                    & 0.94                         & 1.54                         & 13.30                    & 2.14                        \\
DenseTNT \cite{DenseTnT}                                                        & 1.68                         & {\ul 3.63}       & 58.43                    & 0.88                         & 1.28                         & 12.58                    & 1.97                        \\
PRIME \cite{PRIME}                                                              & 1.91                         & 3.82                         & 58.67                    & 1.22                         & 1.55                         & 11.50                    & 2.09                        \\
WIMP \cite{WIMP}                                                                & 1.82                         & 4.03                         & 62.88                    & 0.90                         & 1.42                         & 16.69                    & 2.11                        \\
TPCN \cite{TPCN}                                                                & 1.66                         & 3.69                         & 58.80                    & 0.87                         & 1.38                         & 15.80                    & 1.92                        \\
HOME \cite{HOME}                                                                & 1.70                         & 3.68                         & 57.23                    & 0.89                         & 1.29                         & {\ul \textbf{8.46}}      & {\ul \textbf{1.86}}         \\
mmTransformer \cite{mmtransformer}                                              & 1.77                         & 4.00                         & 61.78                    & 0.87                         & 1.34                         & 15.40                    & 2.03                        \\
MultiModalTransformer \cite{Waypoint}                                           & 1.74                         & 3.90                         & 60.23                    & {\ul 0.84}       & 1.29                         & 14.29                    & 1.94                        \\
LatentVariableTransformer \cite{LVT}                                            & -                            & -                            & -                        & 0.89                         & 1.41                         & 16.00                    & -                           \\
SceneTransformer \cite{SceneTransformer}                                        & 1.81                         & 4.06                         & 59.21                    & {\ul \textbf{0.80}}          & {\ul \textbf{1.23}}          & 12.55                    & {\ul 1.88}      \\ \midrule
\begin{tabular}[c]{@{}c@{}}SSL-Lanes (Ours) \end{tabular} & {\ul \textbf{1.63}}          & {\ul \textbf{3.56}}          & {\ul \textbf{56.71}}     & {\ul 0.84}       & {\ul 1.25}       & 13.26                    & 1.94                        \\ \bottomrule
\end{tabular}%
}
\caption{Comparison of our (best) proposed model and top approaches on the Argoverse Test. The best results are in bold and underlined, and the second best is also underlined.}
\label{tab:sota}
\vspace{-0.25cm}
\end{table*}
\subsubsection{Maneuver Classification}
\quad \textbf{Motivation:} The Lane-Masking and Distance to Intersection pretext tasks are both based on extracting feature and topology information from a HD-map. However, pretext tasks can also be constructed from the overall forecasting task itself. Thus we propose to obtain free pseudo-labels in the form of a `maneuver' the agent-of-interest intends to execute, and define a set of `intentions' to represent common semantic modes (e.g. change lane, speed up, slow down, turn-right, turn-left etc.) We call this pretext task \textit{Maneuver Classification}, and we expect it to provide prior regularization to $\boldsymbol{\mathrm{\Psi}} = \{ {g_{\text{enc}}, \boldsymbol{\mathrm{\Theta}}, \boldsymbol{\Lambda}}\}$, based on driving modes.
\par \textbf{Formulation:} We aim to construct pseudo label to divide agents into different clusters according to their driving behavior and explore unsupervised clustering algorithms to acquire the maneuver for each agent. We find that using naive $k$-Means (on agent end-points) or DBSCAN (on Hausdorff distance between entire trajectories \cite{SSL-Planning}) leads to noisy clustering. We find that constrained $k$-means \cite{ConstrainedKMeans} on agent end-points works best to divide trajectory samples into $C$ clusters equally. We define $C = \{\text{maintain-speed}, \text{accelerate}, \text{decelerate}, \text{turn-left}, \text{turn-right}, \\ \text{lane-change}\}$ and the clustering function as $\rho$. If $p_{\boldsymbol{\mathrm{\Theta}}_{\text{ss}}}(f_{\boldsymbol{\mathrm{\Psi}}}(\mathcal{P}_i, \boldsymbol{X}, \boldsymbol{A}_f))$ is the prediction of agent $i$'s intention and $E_i=(x_{i,\text{GT}}^T, y_{i,\text{GT}}^T)$ is its ground-truth end-point, then the learning objective is to classify each agent maneuver into its corresponding cluster using cross-entropy loss $\mathcal{L}_{ce}$ as:
\begin{equation} \label{eq:8}
\boldsymbol{\mathrm{\Psi}}^{\star}, \boldsymbol{\mathrm{\Theta}}_{\text{ss}}^{\star} = 
\mathop{\text{arg min}}_{\boldsymbol{\mathrm{\Psi}}, \boldsymbol{\mathrm{\Theta}}_{\text{ss}}} \mathcal{L}_{\text{ce}}\Big(p_{\boldsymbol{\mathrm{\Theta}}_{\text{ss}}}(f_{\boldsymbol{\mathrm{\Psi}}}(\mathcal{P}_i, \boldsymbol{X}, \boldsymbol{A}_f)), \rho(E_i)\Big)
\end{equation}
\par \textbf{Benefit of Maneuver Classification Task:} We hypothesize if one can identify the intention of a driver, the future motion of the vehicle will match that maneuver, thereby reducing the set of possible end-points for the agent. We also expect that agents with similar maneuvers will tend to have consistent semantic representations. 
\vspace{-0.4cm}
\subsubsection{Forecasting Success/Failure Classification}
\quad \textbf{Motivation:} In contrast to maneuver classification, which provides coarse-grained prediction of the future, self-supervision mechanisms can also offer a strong learning signal through goal-reaching tasks which are generated from the agent’s trajectories. We propose a pretext task called \textit{Success/Failure Classification}, which trains an agent specialized at achieving end-point goals which directly lead to the forecasting-task solution. We expect this to constrain $\boldsymbol{\mathrm{\Psi}} = \{ {g_{\text{enc}}, \boldsymbol{\mathrm{\Theta}}, \boldsymbol{\Lambda}}\}$ to predict trajectories $\epsilon$ distance away from the correct final end-point. Conceptually, the more examples of successful goal states we collect, the better understanding of the target goal of the forecasting task we have.
\par \textbf{Formulation:} Similar to maneuver classification, we wish to create pseudo-labels for our data samples. We label trajectory predictions as successful ($c=1$) if the final prediction $(x_{i,\text{pred}}^T, y_{i,\text{pred}}^T)$ is within $\epsilon < 2\metre$ of the final end-point $E_i$, and as failure ($c=0$) otherwise. We choose $2\metre$ as our $\epsilon$ threshold because it is also used for miss-rate calculation (\cref{sec:experiments}). In this case, $c \in C = \{0, 1\}$ is the pseudo-label which belongs to label set $C$. If the pretext decoder predicts agent $i$'s final-endpoint as $p_{\boldsymbol{\mathrm{\Theta}}_{\text{ss}}}(f_{\boldsymbol{\mathrm{\Psi}}}(\mathcal{P}_i, \boldsymbol{X}, \boldsymbol{A}_f))$, and given ground-truth end-point $E_i$ its success or failure label is $c_i$, then the pretext loss can be formulated as:
\begin{equation} \label{eq:9}
\boldsymbol{\mathrm{\Psi}}^{\star}, \boldsymbol{\mathrm{\Theta}}_{\text{ss}}^{\star} = 
\mathop{\text{arg min}}_{\boldsymbol{\mathrm{\Psi}}, \boldsymbol{\mathrm{\Theta}}_{\text{ss}}} \mathcal{L}_{\text{ce}}\Big(p_{\boldsymbol{\mathrm{\Theta}}_{\text{ss}}}(f_{\boldsymbol{\mathrm{\Psi}}}(\mathcal{P}_i, \boldsymbol{X}, \boldsymbol{A}_f)), c_i\Big)
\end{equation}
\par \textbf{Benefit of Success/Failure Classification Task:} We hypothesize that this task will especially provide stronger gains for cases where the final end-point is not aligned with the general direction of agent movement for majority of samples given in the dataset, and is thus not well captured by average displacement based supervised loss functions. 
\begin{figure*}
    \centering
    \vspace{0.2cm}
   \includegraphics[width=\textwidth]{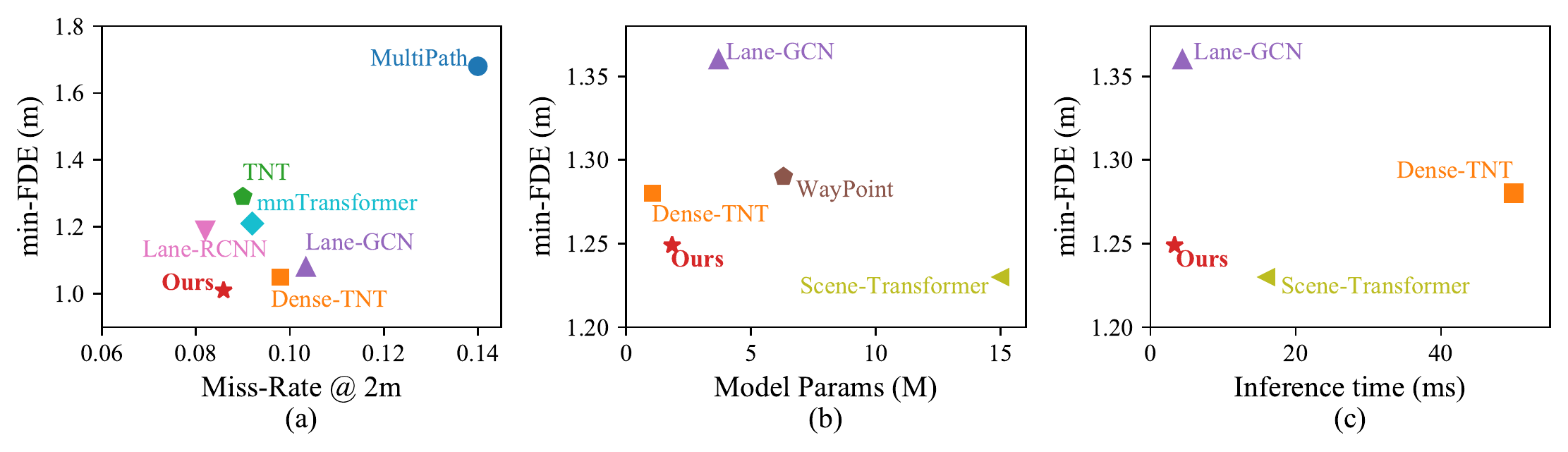}
   \caption{\textit{Left:} $\text{min-FDE}_6$ - $\text{Miss-Rate}_6$ trade-off on Argoverse Validation. Lower-left is better. We optimize both successfully in comparison to other popular approaches. \textit{Right:} We plot min-FDE on Argoverse Test Set against number of model parameters (in millions) and inference time (in milli-seconds). We find that there is a trade-off between min-FDE performance, architectural complexity (as measured by number of parameters) and computational efficiency (as measured by inference time). Our work achieves the best trade-off (lower-left).}
   \label{fig:comparewithsota}
   \vspace{-0.3cm}
\end{figure*}
\subsection{Learning}
As all the modules are differentiable, we can train the model in an end-to-end way. We use the sum of classification, regression and self-supervised losses to train the model.  Specifically, we use:
\begin{equation} \label{eq:10}
\mathcal{L} = \mathcal{L}_{\text{cls}} + \mathcal{L}_{\text{reg}} + \mathcal{L}_{\text{terminal}} + \mathcal{L}_{\text{ss}}    
\end{equation}
For classification and regression loss design, we adopt the formulation proposed in \cite{LaneGCN}. $\mathcal{L}_{\text{terminal}} = \frac{1}{N} \sum_{i=1}^{N}L2\Big((x_{i,\text{pred}}^T, y_{i,\text{pred}}^T), (x_{i,\text{GT}}^T, y_{i, \text{GT}}^T)\Big)$ is a simple L2 loss that minimizes the distance between predicted final-endpoints and the ground-truth. This is because $\mathcal{L}_{\text{reg}}$ is averaged across all time-points $1:T$, and from a practical end user perspective, minimizing the endpoint loss is much more important than weighting loss from all time-steps equally. Our proposed pretext tasks contributes to $\mathcal{L}_{\text{ss}}$. During evaluation, we study each pretext task separately, and their corresponding loss formulations defined in \cref{eq:6}, \cref{eq:7}, \cref{eq:8}, \cref{eq:9} are used for joint training.

\section{Experiments}
\label{sec:experiments}
\par \textbf{Dataset:} Argoverse provides a large-scale dataset \cite{Argoverse} for the purpose of training, validating and testing models, where the task is to forecast 3 seconds of future motions, given 2 seconds of past observations. This dataset has more than 30K real-world driving sequences collected in Miami (MIA)
and Pittsburgh (PIT). Those sequences are further split into train, validation, and test sets, without any geographical overlap. Each of them has 205,942, 39,472, and 78,143 sequences respectively. In particular, each sequence contains the positions of all actors in a scene within the past 2 seconds history, annotated at 10Hz. It also specifies one actor of interest in the scene, with type ‘agent’, whose future 3 seconds of motion are used for the evaluation. The train and validation splits additionally provide future locations of all actors within 3 second horizon labeled at 10Hz, while annotations for test sequences are withheld from the public and used for the leaderboard evaluation. HD map information is available for all sequences.

We have two main requirements for the dataset: (a) \textbf{Scale of Data:} Modern motion forecasting methods and self-supervised learning systems require a large amount of training data to imitate human maneuvers in complex real-world scenarios. Thus, the dataset should be \textit{large-scale} and \textit{diverse}, such that it has a wide range of behaviors and trajectory shapes across different geometries represented in the data. (b) \textbf{Interesting Scenarios for Forecasting Evaluation:} The dataset should be collected for interesting behaviours by biasing sampling towards complex observed behaviours (e.g., lane changes, turns) and road features (e.g., intersections), since we wish to focus on these cases. We find that on the basis of these requirements, as well as its popularity in the the motion forecasting community, Argoverse \cite{Argoverse} is the best candidate to showcase our method. Please refer to the supplementary for more details regarding why we choose to focus on it in comparison to other motion forecasting benchmarks.
\par \textbf{Metrics:} 
ADE is defined as the average displacement error between ground-truth trajectories and predicted trajectories over all time steps. FDE is defined as displacement error between ground-truth trajectories and predicted trajectories at the final time step. We compute $K$ likely trajectories for each scenario with the ground truth label, where $K = 1$ and $K = 6$ are used. Therefore, minADE and minFDE are minimum ADE and FDE over the top $K$ predictions, respectively. Miss rate (MR) is defined as the percentage of the best-predicted trajectories whose FDE is within a threshold (2\,m). Brier-minFDE is the minFDE plus $(1 − p)^2$, where $p$ is the corresponding trajectory probability. 

\par \textbf{Experimental Details:} To normalize the data, we translate and rotate the coordinate system of each sequence so that the origin is at current position $t = 0$ of ‘agent’ actor and x-axis is aligned with its current direction, i.e., orientation from the agent location at $t = −1$ to the agent location at $t = 0$ is the positive x axis. We use all actors and lanes whose distance from the agent is smaller than 100 meters as the input.  We train the model on 4 TITAN-X GPUs using a batch size of 128 with the Adam \cite{Adam} optimizer with an initial learning rate of $1 × 10^{−3}$, which is decayed to $1 × 10^{−4}$ at 100,000 steps. The training process finishes at 128,000 steps and takes about 10 hours to complete. For our final test-set submission, we use success/failure classification as the pretext task, and initialize the map-encoder with the parameters from a model trained with the lane-masking pretext task. To avoid overfitting to the general
directions that agents move, we augment the data from each scene for the test-set submission. We rotate all trajectories in a scene around the scene’s origin by $\gamma$, where $\gamma$ varies from 0$\degree$ to 360$\degree$ in 30$\degree$ intervals. We provide more implementation details in the supplementary.
\vspace{-0.1cm}
\section{Results}
\label{sec:results}
\subsection{Ablation Studies}
\textbf{Effectiveness of Pretext tasks:} We first examine the effect of incorporating our proposed pretext tasks (\cref{sec:method}) with the standard data-driven motion forecasting baseline (\cref{sec:background}). While evaluating the importance of our proposed pretext tasks, we wish to underline that motion prediction for autonomous driving is a safety-critical task, especially at intersections where most of our data is collected, and most accidents also happen. We thus posit that in this situation, even a small error in predicting final locations (FDE) for a given agent can lead to dangerous potential collisions.
\par Results in \cref{tab:ablation} show that all proposed pretext tasks improve motion forecasting performance for Argoverse. Specifically, the lane-masking pretext task improves min-FDE by 8.9\% and MR@2m by 20.3\%. distance to intersection improves min-FDE by 7.1\% and 19.3\%. Maneuver classification improves min-FDE by 6.3\% and MR@2m by 15.4\%. We expect that improving the quality of clustering for maneuvers and thus creating better pseudo-labels will improve this further. Finally, success/failure classification improves min-FDE by 9.8\% and MR@2m by 22.4\%. Moreover, since pretext tasks are not used for inference and only for training, they also do not add any extra parameters or FLOPs to the baseline, thereby increasing accuracy but at no cost to computational efficiency or architectural complexity. We present qualitative results with the different pretext tasks on several hard cases in \cref{fig:qualitative}.
\begin{figure}[t]
  \centering
   \includegraphics[width=\linewidth]{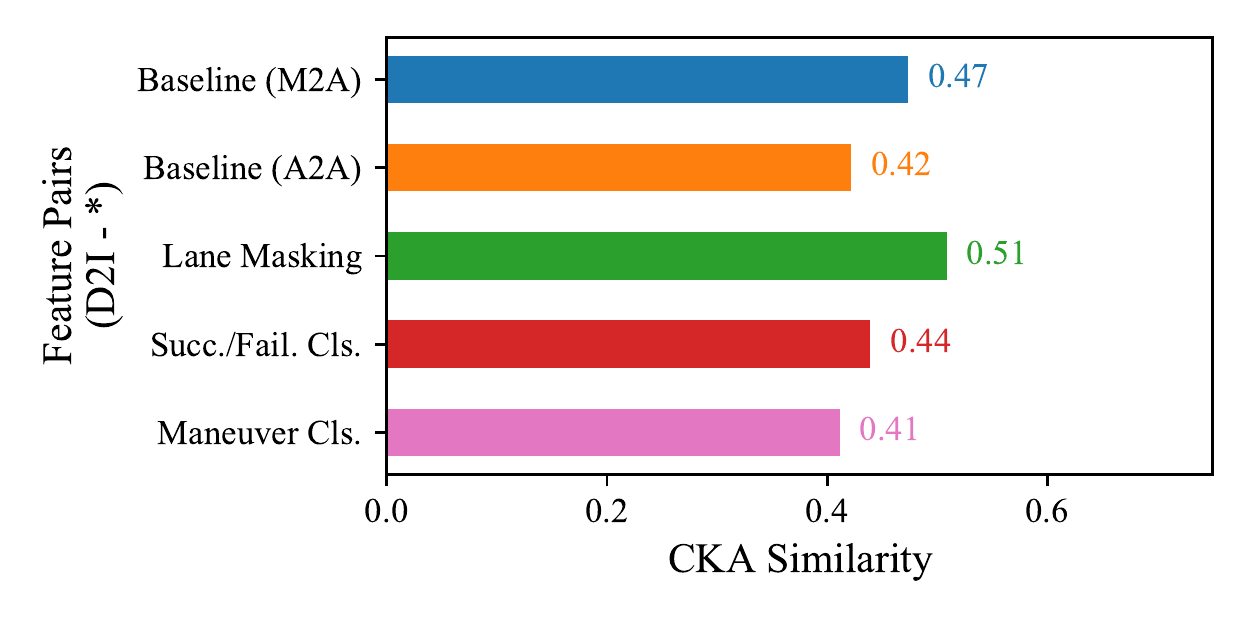}
   \caption{CKA Feature similarity between feature pairs of baseline and different pretext tasks. Similarity score is 1 for completely overlapping features and 0 for completely divergent features.}
   \label{fig:cca-similarity}
   \vspace{-0.2in}
\end{figure}
\par \textbf{Similarity in feature space:} We analyze the CKA similarity \cite{CCA} between the representations learnt by: a model trained with pretext task `D2I' (refers to distance to intersection task) and baseline; two models trained with different pretext tasks. In \cref{fig:cca-similarity}, Base(M2A) refers to $\Tilde{\boldsymbol{p}}_i$, Base(A2A) refers to $\acute{\boldsymbol{p}}_i$ (see \cref{eq:3}), `Mask' refers to lane-masking, `success/fail' refers to success or failure classification task and `intention' suggests maneuver classification. 
\par Our main questions are: (a) how much does the pretext task feature differ from the baseline? (b) do the features from different pretext tasks collapse to the same feature? First we note that representation learned by D2I does not \textit{collapse} to the same representation learned by Mask or Success/Fail or Intention. Secondly we note that D2I features are quite different from Base-M2A features $\Tilde{\boldsymbol{p}}_i$ and Base-A2A features $\acute{\boldsymbol{p}}_i$, which suggests that \textit{task-specific regularization} has indeed resulted in different parameters.

\begin{table*}[]
\small
\centering
\resizebox{\textwidth}{!}{%
\begin{tabular}{@{}c|cc|c|ccc@{}}
\toprule
\multirow{2}{*}{\textbf{Description}}                                                       & \multicolumn{2}{c|}{\textbf{Experimental Setup}}                                                                                                                                                   & \multirow{2}{*}{\textbf{Method}} & \multicolumn{1}{c}{\multirow{2}{*}{\textbf{$\textrm{minADE}_6$}}} & \multicolumn{1}{c}{\multirow{2}{*}{\textbf{$\textrm{minFDE}_6$}}} & \multicolumn{1}{c}{\multirow{2}{*}{\textbf{$\textrm{MR}_6$}}} \\
                                                                                            & \textbf{Training}                                                                            & \textbf{Validation}                                                                                           &                                  & \multicolumn{1}{c}{}                                              & \multicolumn{1}{c}{}                                              & \multicolumn{1}{c}{}                                          \\ \midrule
\multirow{2}{*}{\begin{tabular}[c]{@{}c@{}}Effects of limited\\ training data\end{tabular}} & \multirow{2}{*}{25\% of train}                                                            & \multirow{2}{*}{All}                                                                                   & Baseline                         & 0.82                                                              & 1.33                                                              & 14.66                                                         \\
                                                                                            &                                                                                           &                                                                                                        & Ours                             & {\ul \textbf{0.78}}                                               & {\ul \textbf{1.22}}                                               & {\ul \textbf{12.63}}                                          \\ \midrule
\multirow{2}{*}{\begin{tabular}[c]{@{}c@{}}Effects of\\ new domain\end{tabular}}            & \multirow{2}{*}{\begin{tabular}[c]{@{}c@{}}100\% PIT +\\ 20\% MIA\end{tabular}}           & \multirow{2}{*}{MIA val}                                                                               & Baseline                         & 0.88                                                              & 1.46                                                              & 17.21                                                         \\
                                                                                            &                                                                                           &                                                                                                        & Ours                             & {\ul \textbf{0.85}}                                               & {\ul \textbf{1.34}}                                               & {\ul \textbf{14.96}}                                          \\ \midrule
\multirow{2}{*}{\begin{tabular}[c]{@{}c@{}}Performance on\\ difficult maneuvers\end{tabular}} & \multirow{2}{*}{All}                                                                      & \multirow{2}{*}{\begin{tabular}[c]{@{}c@{}}Turning \&\\ lane changing\end{tabular}}                    & Baseline                         & 0.90                                                              & 1.53                                                              & 19.90                                                         \\
                                                                                            &                                                                                           &                                                                                                        & Ours                             & {\ul \textbf{0.84}}                                               & {\ul \textbf{1.34}}                                               & {\ul \textbf{14.93}}                                          \\ \midrule
\multirow{2}{*}{\begin{tabular}[c]{@{}c@{}}Effects of\\ imbalanced data \end{tabular}} & \multirow{2}{*}{\begin{tabular}[c]{@{}c@{}}2x straight\\ 1x other maneuvers\end{tabular}} & \multirow{2}{*}{\begin{tabular}[c]{@{}c@{}}Turning \&\\ lane changing\end{tabular}}                    & Baseline                         & 0.94                                                              & 1.65                                                              & 21.53                                                         \\
                                                                                            &                                                                                           &                                                                                                        & Ours                             & {\ul \textbf{0.90}}                                               & {\ul \textbf{1.49}}                                               & {\ul \textbf{17.97}}                                          \\ \midrule
\multirow{2}{*}{\begin{tabular}[c]{@{}c@{}}Effects of\\ noisy data\end{tabular}}            & \multirow{2}{*}{All}                                                                      & \multirow{2}{*}{\begin{tabular}[c]{@{}c@{}}Gaussian noise ($\sigma=0.2$)\\ with $p=0.25$\end{tabular}} & Baseline                         & 1.01                                                              & 1.37                                                              & 15.59                                                         \\
                                                                                            &                                                                                           &                                                                                                        & Ours                             & {\ul \textbf{0.96}}                                               & {\ul \textbf{1.24}}                                               & {\ul \textbf{11.98}}                                          \\ \midrule
\multirow{2}{*}{\begin{tabular}[c]{@{}c@{}}Effects of\\ noisy data\end{tabular}}            & \multirow{2}{*}{All}                                                                      & \multirow{2}{*}{\begin{tabular}[c]{@{}c@{}}Gaussian noise ($\sigma=0.2$)\\ with $p=0.5$\end{tabular}} & Baseline                         & 1.19                                                              & 1.56                                                              & 20.64                                                         \\
                                                                                            &                                                                                           &                                                                                                        & Ours                             & {\ul \textbf{1.13}}                                               & {\ul \textbf{1.40}}                                               & {\ul \textbf{15.65}}                                          \\ \bottomrule
\end{tabular}%
}
\vspace{0.05in}
\caption{Different experimental settings to provide evidence for why SSL-based training helps motion forecasting}
\label{tab:analysis}
\vspace{-0.2in}
\end{table*}
\subsection{Comparison with State-of-the-Art}
\textbf{Performance:} We compare our approach with top entries on Argoverse
motion forecasting leaderboard \cite{Argoverse} in \cref{tab:sota}. SSL-Lanes improves the metrics for $K = 1$ convincingly and outperforms existing approaches w.r.t. $\text{min-ADE}_1$, $\text{min-FDE}_1$ and $\text{MR}_1$. We are also strongly competitive w.r.t. $\text{min-ADE}_6$, $\text{min-FDE}_6$ and $\text{MR}_6$ against top approaches, with a relatively simple architecture.
\par \textbf{Trade-off between min-FDE and Miss-Rate:} $\text{min-FDE}_6$ and $\text{MR}_6$ are both important for autonomous robots to optimize. Ideally we wish for both of these metrics to be low. However, there exists a frequent trade-off between them.  We compare this trade-off in \cref{fig:comparewithsota}(a) with six other popular motion forecasting models (in terms of citations and GitHub stars), namely: Lane-GCN \cite{LaneGCN}, Lane-RCNN \cite{LaneGCN}, MultiPath \cite{Multipath}, mm-Transformer \cite{mmtransformer}, TNT \cite{TNT} and Dense-TNT \cite{DenseTnT} on the Argoverse validation set. We are on the lowest-left of meaning we optimize both $\text{min-FDE}_6$ and $\text{MR}_6$ successfully in comparison to other top models.
\par \textbf{Trade-off between accuracy, efficiency and complexity:} We are the first to point out a trade-off that exists for current state-of-the-art motion forecasting models between forecasting performance, architectural complexity and inference speed. This is illustrated in  \cref{fig:comparewithsota}(b)-(c). NN+Map \cite{Argoverse} (see \cref{tab:sota}) is a simple nearest-neighbor based approach that also uses map-features, and while it has advantages in terms of fast inference and low model complexity, the forecasting performance is very low. MultiPath \cite{Multipath} is a very popular approach that has reasonable accuracy and inference speed but is parametrically heavy due to it's use of convolutional kernels. Lane-GCN is a vector based approach \cite{LaneGCN} has comparatively fast inference time and high accuracy, but uses multiple GNN layers which can lead to problems with over-smoothing for map-encoders \cite{Oversmoothing} and also has a complicated four-stage fusion mechanism. Lane-RCNN \cite{Lanercnn} proposes to capture interactions between agents and map using not just a single vector, but a local interaction graph per agent - this adds huge number of hyper-parameters to the model and makes it very complex. Transformer-based models \cite{mmtransformer, Waypoint, SceneTransformer, LVT} also suffer in this regard. Scene-transformer for example has 15M parameters and uses heavy augmentation to prevent overfitting. A light high-performing model is Dense-TNT \cite{DenseTnT}. However, Dense-TNT's inference speed on average is 50ms per agent, because it proposes a time-intensive optimization algorithm to find a dense goal set that minimizes the expected error of the given set. In contrast to these popular models, our approach has high accuracy (min-FDE: 1.25m, MR: 13.3\%) while also having low architectural complexity (1.84M parameters) and high inference speed (3.30 ms). Thus it provides a great balance for application to real-time safety-critical autonomous robots.

\vspace{-0.1cm}
\par \textbf{Qualitative Results:} 
\label{qualitative}
We present some multi-modal prediction trajectories on several hard cases shown in \cref{fig:qualitative}. The yellow trajectory represents the observed 2s. Red represents ground truth for the next 3s and green represents the multiple forecasted trajectories for those 3s. In Row 1, the agent turns right at the intersection. The baseline misses this mode completely, despite having access to the map. The model trained with lane-masking successfully predicts this right turn within 2m of the ground-truth end-point. In Row 2, the agent has a noisy past history and accelerates while turning left at the intersection. The pretext task distance-to-intersection can correctly capture this, while the baseline has only one trajectory covering this mode but vastly overshoots the ground-truth. Interestingly, we note that the success/failure pretext task is unable to capture this mode. We believe this is due to a stronger prior imposed by the model during learning. In Row 3, we have an agent accelerating while going straight at an intersection. We find that the maneuver classification pretext task is the only model that correctly predicts trajectories aligned with the ground-truth. In Row 4, we have an agent turning left at an intersection. Most of the predictions of other models predicts that the agent will go straight. The success/failure pretext task however picks up on the left-turn, possibly due to the priors imposed upon it by end-point conditioning. 
\\ Overall, SSL-Lanes can capture left and right turns better, while also being able to discern acceleration at intersections. Our pretext tasks provide priors for the model and provides data-regularization for free. We believe this can improve forecasting through better understanding of map topology, agent context with respect to the map, and generalization with respect to imbalance implicitly present in data.
\subsection{When does SSL help Motion Forecasting?}
\textbf{Hypotheses:} We hypothesize that training with SSL pretext tasks probably helps motion forecasting as following: (a) Topology-based context prediction assumes feature similarity or smoothness in small neighborhoods of maps. Such a context-based feature representation can greatly improve prediction performance, especially when the neighborhoods are small. (b) Clustering and classification assumes that feature similarity implies target-label similarity and can group distant nodes with similar features together, leading to better generalization. (c) Supervised learning with imbalanced datasets sees significant degradation in performance. Although most of the data samples in Argoverse are at an intersection, a significantly large number involve driving straight while maintaining speed. Recent studies \cite{SSL-Imbalance} have shown that SSL tends to learn richer features from more frequent classes which also allows it to generalize to to other classes better. 
\\ \textbf{Experiments:} In order to provide evidence for our hypotheses, we propose to design 6 different training and testing setups as shown in \cref{tab:analysis}. We use success/failure classification as the pretext task, and all models are trained for 50,000 steps. We initialize the map-encoder with the parameters from a model trained with the lane-masking pretext task. \\
Our \textit{first} setting is to train with 25\% of the total data available for training and testing on the full validation set. We expect the SSL-based task to capture richer features and generalize better than the baseline. Our \textit{second} setting assumes that SSL also generalizes to topology from different cities and trains on 100\% of data from Pittsburgh (PIT) but only 20\% of data from Miami (MIA). For evaluation, we only test on data examples taken from the city of MIA. For our \textit{third} setting, we assume that SSL learns superior features and can thus perform better in difficult cases like lane-changes and turning cases. For evaluation, we only test on data examples which involves these difficult cases. In our \textit{fourth} setting, we choose to explicitly train with data that contains 2$\times$ `straight-with-same-speed' maneuver and 1$\times$ all other maneuvers. We test only on lane-changes and turning cases from validation. Finally in order to test the effect of noise on motion forecasting performance, we take two models already trained on full data. We now take the full validation set, randomly select agent trajectories or map nodes with probability $p=0.25$ and $p=0.5$, and then add Gaussian noise with zero mean and $0.2$ variance to their features. We expect this to have the most impact on forecasting performance as compared to all other settings since this is the most aggressive form of corruption. But we also expect SSL-based pretext task training to provide robustness to noise for free due to better generalization capabilities. Our takeaway from these experiments is that there is strong evidence SSL-based tasks do provide better generalization capabilities and can thus prove to be more effective than pure supervised training based approaches.

\vspace{-0.1cm}
\section{Conclusion}
\label{sec:conclusion}
We propose SSL-Lanes to leverage supervisory signals generated from data for free in the form of pseudo-labels and integrate it with a standard motion forecasting model. We design four pretext tasks that can take advantage of map-structure and similarities between agent dynamics to generate these pseudo-labels, namely: lane masking, distance to intersection prediction, maneuver classification and success/failure classification. We validate our proposed approach by achieving competitive results on the challenging large-scale Argoverse benchmark. The main advantage of SSL-Lanes is that it has high accuracy combined with low architectural complexity and high inference speed. We further demonstrate that each proposed SSL pretext task improves upon the baseline, especially in difficult cases like left/right turns and acceleration/deceleration. We also provide hypotheses and experiments on why SSL-Lanes can improve motion forecasting. 
\par \textbf{Limitations:} A limitation of our framework is that it uses the different losses for our formulation only in a 1:1 ratio without tuning them. We also use only one pretext task at a time and do not explore the combination of these different tasks. For our future work, we plan to incorporate meta-learning \cite{MetalearningSSL} to identify an effective combination of pretext tasks and automatically balance them---we expect that this will lead to more gains in terms of forecasting performance. Another limitation is that we report improvements with SSL-pretext tasks in scenarios without specifically considering multiple heavily interacting agents. In the future we would like to explore how the interactions between road agents can influence our SSL losses on the interaction split of the Waymo Open Motion dataset (WOMD) \cite{WOMD}. Finally, we explore generalization in terms of implicit data imbalance only in comparison to pure supervised training on the same dataset from which training samples are derived. We would like to study the generalization of our work to other datasets without re-training. 

\section*{Acknowledgements}
\vspace{-0.1cm}
This research was funded by the Mitacs Accelerate Program and Gatik Inc. This article solely reflects the opinions and conclusions of its authors.

{\small
\bibliographystyle{ieee_fullname}
\bibliography{main}
}

\clearpage
\begin{center}
    \textbf{\Large Supplementary Material} 
\end{center}
\setcounter{section}{0}
\renewcommand{\theequation}{\arabic{equation}}
\renewcommand{\thefigure}{\arabic{figure}}
\section{Detailed Network Architecture for Baseline}
\begin{figure*}
    \centering
    \includegraphics[width=0.75\linewidth]{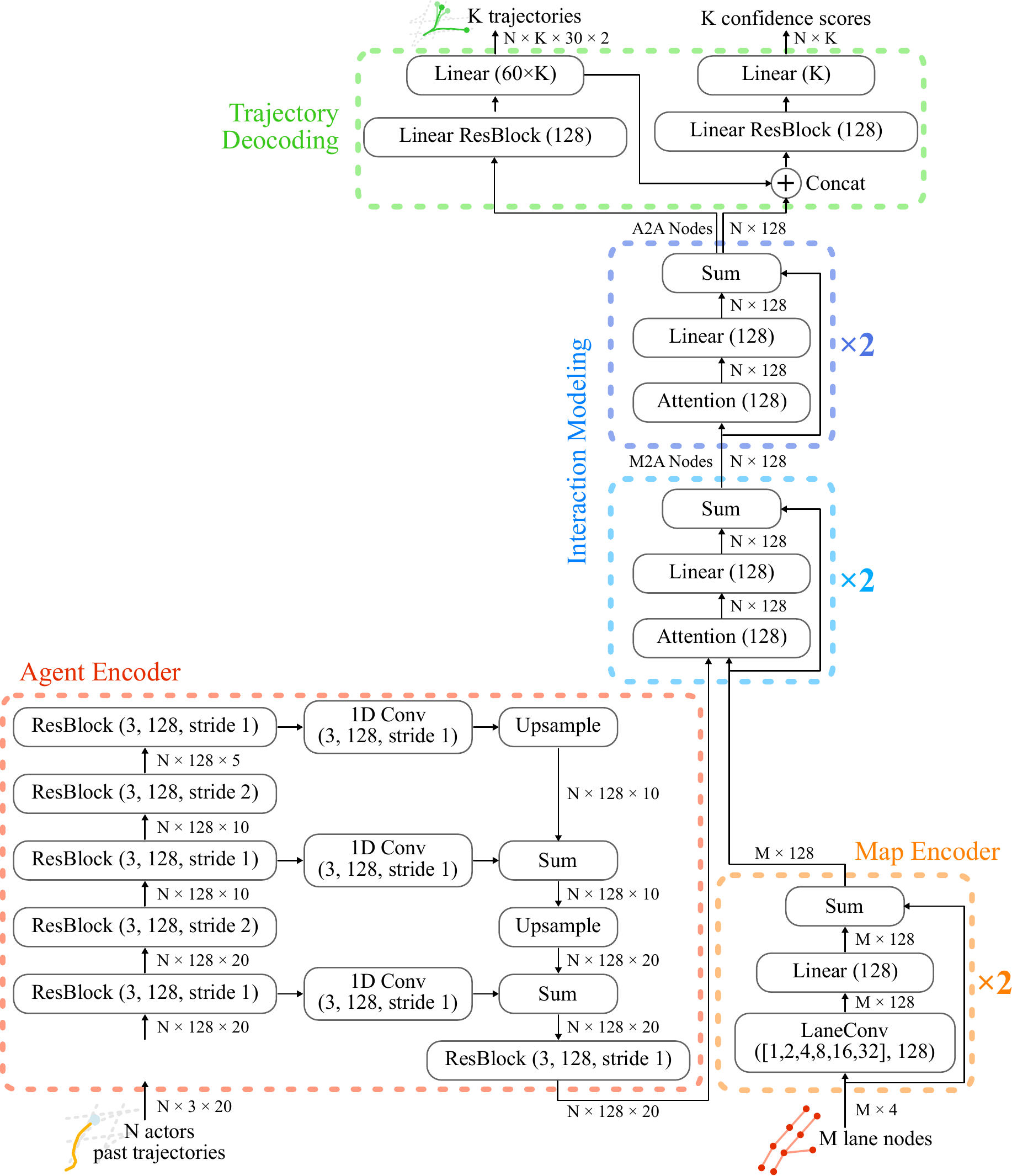}
    \caption{Architecture of the baseline model}
    \label{fig:baseline}
\end{figure*}

We provide the detailed network architecture of our
baseline in \cref{fig:baseline}. 
\\ For the \textit{agent feature extractor}, the architecture is similar to \cite{LaneGCN}. We use an 1D CNN to process the trajectory input. The output is a temporal feature map, whose element at $t = 0$ is used as the agent feature. The network has three groups/scales of 1D convolutions. Each group consists of two residual blocks \cite{residualblock}, with the stride of the first block as 2. Feature Pyramid Network (FPN) \cite{FPN} fuses the multi-scale features, and applies another residual block to obtain the output tensor. For all layers, the convolution kernel size is 3 and the number of output channels is 128. Layer normalization \cite{layernorm} and Rectified Linear Unit (ReLU) are used after each convolution. 
\\ The \textit{map feature extractor} has two LaneConv residual \cite{residualblock} blocks  which are the stack of a LaneConv(1, 2, 4, 8, 16, 32) and a linear layer, as well as a shortcut. All layers have 128 feature channels. Layer normalization \cite{layernorm} and ReLU are used after each LaneConv and linear layer. 
\\ For the map-aware agent feature (M2A) module, the distance threshold is $12$m. It is $100$m for the agent-to-agent (A2A) interaction module.  The two  \textit{interaction modules} have two residual blocks, which consist of a stack of an attention layer and a linear layer, as well as a residual connection. All layers have 128 output feature channels. 
\\ Taking the interaction-aware actor features as input, our \textit{trajectory decoder} is a multi-modal prediction header that outputs the final motion forecasting. For each agent, it predicts $K$ possible future trajectories
and confidence scores. The header has two branches, a regression branch to predict the trajectory of each mode and a classification branch to predict the confidence score of each mode.
\\ \textit{Key differences with Lane-GCN} \cite{LaneGCN}: Our main difference is we use two Lane-Conv blocks instead of four as map-feature extractor in order to prevent over-smoothing in GNNs \cite{Oversmoothing}. We also do not use the four-way fusion proposed by Lane-GCN and do away with the agent to map (A2M) and the map to map (M2M) interaction blocks, which saves compute and memory.

\section{Implementation of Pretext Tasks}
In this section, we discuss various design decisions for the proposed pretext tasks.
\subsection{Lane-Masking}
For this pretext task, we mask $m_a$ percent of every lane and reconstruct its features.  \begin{table}[h]
\centering
\resizebox{0.8\linewidth}{!}{
\begin{tabular}{@{}r|c|ccc@{}}
\toprule
\textbf{Method}                & \textbf{$\textrm{m}_a$} & \textbf{$\textrm{minADE}_6$} & \textbf{$\textrm{minFDE}_6$} & \textbf{$\textrm{MR}_6$} \\ \midrule
Baseline                                          & -  & 0.73                         & 1.12                         & 11.07                    \\ \midrule 
Random Masking & 0.4 & 0.71 & 1.03 & 9.11 \\ \midrule
Lane-Masking & 0.3 & 0.71 & 1.04 & 9.02 \\ \midrule
Lane-Masking & 0.4 & \uline{\textbf{0.70}} & \uline{\textbf{1.02}} & \uline{\textbf{8.84}} \\ \midrule
Lane-Masking & 0.5 & 0.71 & 1.05 & 9.31 \\ \bottomrule
\end{tabular} 
}
\caption{Effect of masking ratio ($m_a$) on forecasting performance for lane-masking task}
\label{tab:lane-masking-ablation}
\vspace{-0.2in}
\end{table}
In \cref{tab:lane-masking-ablation}, we study the influence of masking ratio on the final forecasting performance. Random masking refers to masking out $m_a$ percent random map nodes and lane-masking refers to masking out $m_a$ percent of lanes in the map. We finally choose $m_a=0.4$ as the most effective parameter for the lane-masking pretext task, which outperforms random masking. The model infers missing lane-nodes to produce plausible outputs during reconstruction. We hypothesize that
this reasoning is linked to learning useful representations.
\subsection{Distance to Intersection}
For this pretext task, we explore two different options for framing the problem of predicting the distance to the nearest intersection node in \cref{tab:d2i-ablation}. We first explore predicting this distance as a classification task. We group the lengths into four categories:
$d_{ij} = 1$, $d_{ij} = 2$, $d_{ij} = 3$, $d_{ij} = 4$ and $d_{ij} >= 5$. We however find that this is harder to optimize than the regression loss proposed in \cref{eq:7}, which we finally choose as our loss for the distance to intersection pretext task.
\begin{table}[h]
\centering
\resizebox{\linewidth}{!}{
\begin{tabular}{@{}r|c|ccc@{}}
\toprule
\textbf{Method}                & \textbf{Pretext Loss} & \textbf{$\textrm{minADE}_6$} & \textbf{$\textrm{minFDE}_6$} & \textbf{$\textrm{MR}_6$} \\ \midrule
Baseline                                          & -  & 0.73                         & 1.12                         & 11.07                    \\ \midrule 
Distance to Intersection & Classification & 0.72
& 1.06 & 9.64 \\ \midrule
Distance to Intersection & Regression & \uline{\textbf{0.71}} & \uline{\textbf{1.04}} & \uline{\textbf{8.93}} \\
 \bottomrule
\end{tabular}%
}
\caption{Effect of pretext loss type on forecasting performance for distance to intersection task}
\label{tab:d2i-ablation}
\vspace{-0.2in}
\end{table}
\subsection{Maneuver Classification}
\begin{figure}[h]
  \centering
   \includegraphics[width=0.8\linewidth]{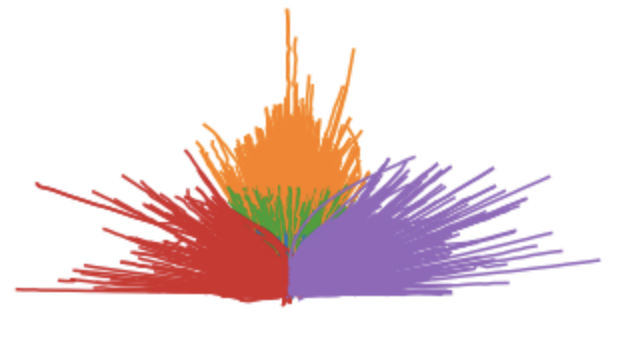}
   \caption{Modes of driving from unsupervised clustering of data}
   \label{fig:manuver clustering}
   \vspace{-0.25in}
\end{figure}

For this pretext task, we first divide the lateral and longitudinal maneuvers by choosing a threshold angle of $20\degree$ from the vertical. We next find that constrained k-means \cite{ConstrainedKMeans} on agent end-points for lateral and longitudinal maneuvers works best to separate the trajectory samples into different clusters. This is illustrated in \cref{fig:manuver clustering}. For differentiating the longitudinal maneuvers from the lane-change maneuver, we check a combination of the distance from the lane centerlines for start and stop positions and the orientations of the nearest centerline for start and stop positions.
\vspace{-0.18in}
\subsection{Success/Failure Classification}
For this pretext task, the primary bottleneck is the fact that the number of positive examples if far fewer than the number of negative examples. This is because there are only a few success examples in a $2$m area near the end-point of a single recorded ground-truth trajectory, while the rest of the points in the scene can be considered as failure examples. We consider first setting $\epsilon=3$m, i.e. a wider area for success examples, and then reducing it to $\epsilon=2$m linearly over the total number of training steps. We find that this can actually harm the final forecasting performance. We thus follow \cite{objectsaspoints} to use focal loss to train our auxiliary classification task. 
\section{Qualitative Results}

\begin{figure*}[ht]
    \centering
    \includegraphics[width=\linewidth]{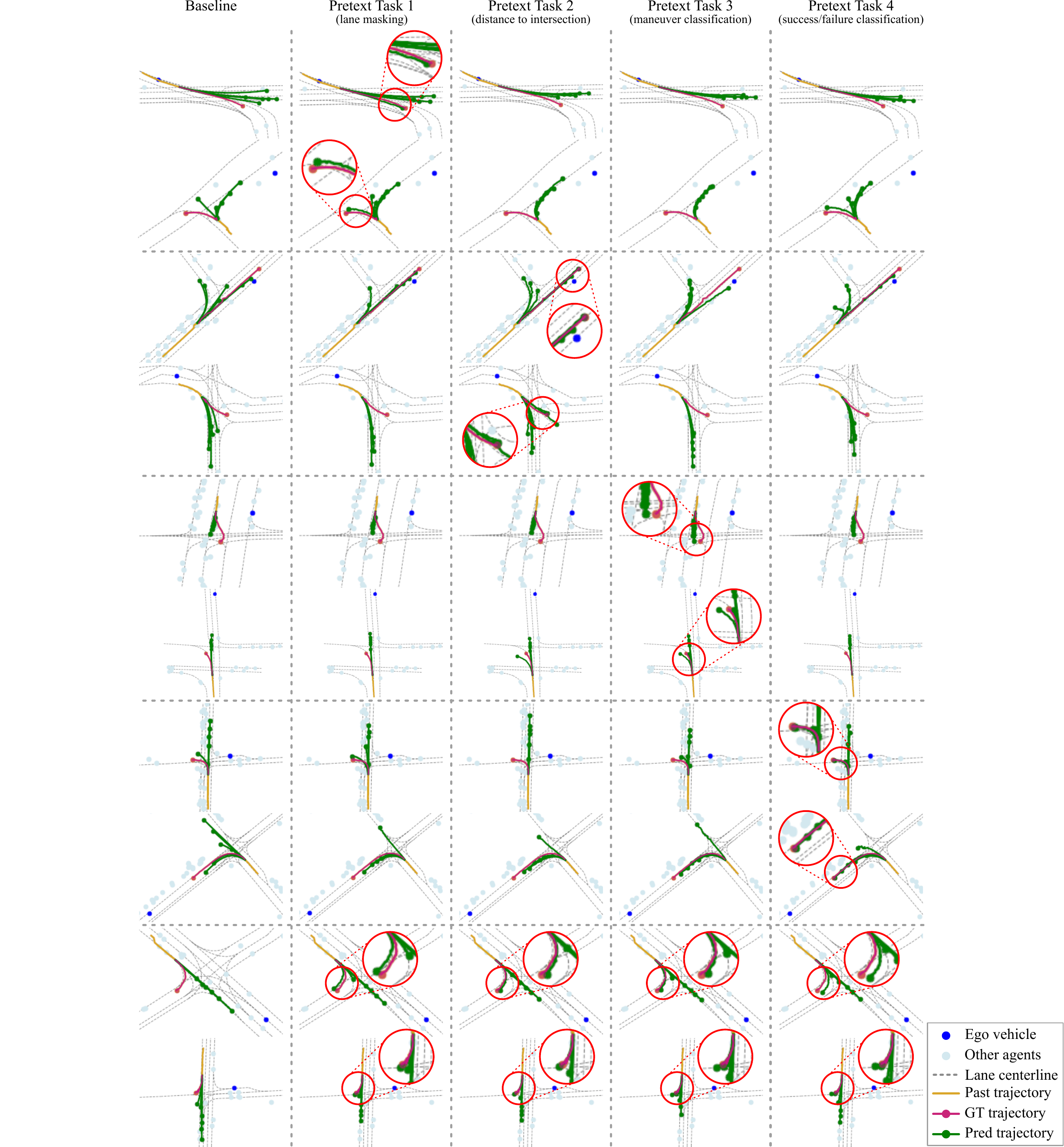}
   \caption{Qualitative results for our proposed SSL-Lanes pretext tasks on the Argoverse \cite{Argoverse} validation set. We outperform the baseline on several difficult cases at intersections and lane-changes.}
   \label{fig:supp-qualitative-all}
\end{figure*}

We first present some multi-modal prediction trajectories on several hard cases shown in \cref{fig:qualitative}. SSL-Lanes can capture left and right turns better, while also being able to discern acceleration at intersections. Our pretext tasks provide priors for the model and provides data-driven regularization for free. This can improve forecasting because of better understanding of map topology, agent context with respect to the map, and also improve generalization for maneuver imbalance implicitly present in data. We next provide more visual results of our proposed SSL-Lanes on the Argoverse validation set in \cref{fig:supp-qualitative-all}. Generally, these qualitative results demonstrate the effectiveness of our proposed pretext tasks.

\section{Discussion: SSL-Lanes vs. State-of-the-Art}
We use this section to distinguish our work from methods that we believe have similar intuition but very different construction, in order to highlight its novelty and value.

\begin{itemize}
    \item SSL-Lanes vs. VectorNet \cite{Vectornet}: Vector-Net is the only other motion forecasting work that proposes to randomly mask out the input node features belonging to either scene context or agent trajectories, and ask the model to reconstruct the masked features. Their intuition is to encourage the graph networks to better capture the interactions between agent dynamics and scene context. However, our motivation differs from VectorNet in two respects: (a) We propose to use masking to learn local map-structure better, as opposed to learning interactions between map and the agent. This is an easier optimization task, and we out-perform VectorNet. (b) A lane is made up of several nodes. We propose to randomly mask out a certain percentage of each lane. This is a much stronger prior as compared to randomly masking out any node (which may correspond to either a moving agent or map) and ensures that the model pays attention to all parts of the map.
    \item SSL-Lanes vs. CS-LSTM \cite{cs-lstm}: CS-LSTM appends the encoder context vector with a one-hot vector corresponding to the lateral maneuver class and a one-hot vector corresponding to the longitudinal maneuver class. Subsequently, the added maneuver context allows the decoder LSTM to generate maneuver specific probability distributions. This construction however is quite different from our work because it is not auxiliary in nature - it always outputs and appends a maneuver to the decoder, even during inference. This we believe is too strong of a bias for the prediction model, especially given the fact that the maneuvers are generated using very simple velocity profiles and not from careful mining of the data. In our conditioning, the maneuvers are mined from data and the final motion prediction does not depend directly on them. We believe this design is much more flexible since it allows to generate more supervisory signals in the form of maneuvers during training, but at the same time does not require an explicit maneuver to condition the final future forecast trajectory output during inference.  
    \item SSL-Lanes vs. MultiPath \cite{Multipath}: MultiPath is also not auxiliary in nature: it factorizes motion uncertainty into intent uncertainty and control uncertainty; models the uncertainty over a discrete set of intents with a softmax distribution; and then outputs control uncertainty as a Gaussian distribution dependent on each waypoint state of the anchor trajectory (corresponding to the intent). While this construction is highly intuitive and effective by design, it is very different from our SSL-based construction. Ours is an auxiliary task which provides supervision during training, and effectively functions as a regularizer, while being general enough to be used with any other data-driven motion forecasting model.
\end{itemize}
\section{Discussion: Choice of Dataset} 
We now compare the commonly used motion-forecasting datasets, i.e., nuScenes \cite{nuScenes}, Waymo-Open-Motion-Dataset (WOMD) \cite{WOMD} and Argoverse \cite{Argoverse}. We individually discuss why Argoverse is best positioned to bring out the benefits of our proposed work.
\begin{itemize}
    \item \textit{Scale of Data:} We first compare the dataset size and diversity. We note that Argoverse is not only larger and more diverse than nuScenes, but also has greater number of training samples and unique trajectories compared to WOMD. \\
\begin{table}[h]
\small
\centering
\resizebox{0.8\linewidth}{!}{
\begin{tabular}{@{}c | c c c@{}}
\toprule
  & nuScenes & WOMD & Argoverse \\ [0.5ex] 
 \hline\hline
Number of Unique Tracks: & 4.3k & 7.65m & 11.7m \\
\hline 
Number of Training Segments: & 1k & 104k & 324k \\
 \hline 
\end{tabular}
}
\end{table}
\item \textit{Interesting Scenarios for Forecasting Evaluation:} We next compare if the datasets specifically mines for interesting scenarios, which is the area we want to improve the current baseline. nuScenes was not collected to capture a wide diversity of complex and interesting driving scenarios. WOMD on the other hand specifically mines for pairwise interaction scenarios, where the main objective is to improve forecasting for interacting agents. However, the scope of our study is to primarily focus on motion at intersections undergoing lane-changes and turns. We expect the SSL-losses to improve understanding of the context/environment, trajectory embeddings and address data-imbalance w.r.t. maneuvers. We leave heavy interaction-based use cases for future work. Finally, Argoverse mines for interesting motion patterns at intersections, which involve lane-changes, acceleration/deceleration, and turns. We thus find this dataset best suited to showcase our proposed method.
\item \textit{Community focus on Argoverse:} We also find that many popular motion forecasting methods published by the robotics community have also included evaluations only on the Argoverse dataset including: \href{https://arxiv.org/pdf/2007.13732.pdf}{Lane-GCN},  \href{https://arxiv.org/abs/2101.06653}{Lane-RCNN}, \href{https://arxiv.org/abs/2103.04027}{PRIME}, \href{https://arxiv.org/abs/2204.05859}{DCMS},  \href{https://arxiv.org/abs/2103.03067}{TPCN},  \href{https://arxiv.org/pdf/2103.11624.pdf}{mm-Transformer}, \href{https://openaccess.thecvf.com/content/CVPR2022/papers/Zhou_HiVT_Hierarchical_Vector_Transformer_for_Multi-Agent_Motion_Prediction_CVPR_2022_paper.pdf}{HiVT}, \href{https://arxiv.org/abs/2109.06446}{Multi-modal Transformer}, \href{https://arxiv.org/abs/2111.01592}{DSP} etc. This makes it easier for us to position our work with respect to these approaches.
\end{itemize}
\section{Discussion: Potential of this Work}
We expect this work to influence real world deployment of SSL forecasting methods for autonomous driving. Another use case for this work is realistic behavior generation in traffic simulation. The general construction of the prediction problem, inspired by \cite{LaneGCN}, enables a generic understanding of how an object moves in a given environment without memorizing the training data. A neural network may learn to associate particular areas of a scene with certain motion patterns. To prevent this, we centre around the agent of interest and normalize all other trajectory and map coordinates with respect to it. We predict relative motion as opposed to absolute motion for the future trajectory. This helps to learn general motion patterns. Reconstructing the map or predicting distances from map elements are conducted in a frame-of-reference relative to the agent of interest. This helps in learning general map connectivity. Following work in pedestrian trajectory prediction, we also additionally add random rotations to the training trajectories to reduce directional bias. Furthermore, we provide strong evidence that SSL-based tasks provide better generalization compared to pure supervised training, thereby having the ability to effectively reuse the same prediction model across different scenarios. 

\end{document}